\documentclass[11pt,a4paper]{article}
\usepackage[hyperref]{acl2020}
\usepackage{times}
\usepackage{latexsym}

\usepackage{microtype}

\usepackage{amssymb}
\usepackage{graphicx}
\usepackage{subcaption}
\usepackage{url}

\aclfinalcopy 
\def\aclpaperid{276} 

\title{On the Effects of Knowledge-Augmented Data in Word Embeddings}

\author{Diego Ramirez-Echavarria \\
  University College London \\
  \texttt{diego.echavarria.17@ucl.ac.uk} \\ \And
  Antonis Bikakis \\
  University College London \\
  \texttt{a.bikakis@ucl.ac.uk} \\ \AND
  Luke Dickens \\
  University College London \\
  \texttt{l.dickens@ucl.ac.uk} \\ \And
  Rob Miller \\
  University College London \\
  \texttt{r.s.miller@ucl.ac.uk} \\ \And
  Andreas Vlachidis \\
  University College London \\
  \texttt{a.vlachidis@ucl.ac.uk} }

\date{}

\begin{document}
\maketitle
\begin{abstract}
This paper investigates techniques for knowledge injection into word embeddings learned from large corpora of unannotated data. These representations are trained with word co-occurrence statistics and do not commonly exploit syntactic and semantic information from linguistic knowledge bases, which potentially limits their \textit{transferability} to domains with differing language distributions or usages. We propose a novel approach for linguistic knowledge injection through data augmentation to learn word embeddings that enforce semantic relationships from the data, and systematically evaluate the impact it has on the resulting representations. We show our knowledge augmentation approach improves the intrinsic characteristics of the learned embeddings while not significantly altering their results on a downstream text classification task.
\end{abstract}

\section{Introduction}
\label{sec:introduction}

Word embeddings have become pervasive across many NLP tasks. Some of the most common word embedding models are learned from word co-occurrence statistics in text \citep{Mikolov2013b, Pennington2014}. These models are based on the distributional semantics idea that the meaning of a word can be inferred from its context. Despite the widespread adoption of these models, in practice the quality of the learned embeddings is determined by the size, noisiness, and linguistic diversity of the text corpus used to train them. Moreover, the distributional information that can be captured by these models is domain specific, meaning that these word embeddings capture the regularities of a particular use of language. This limits the transferability of the learned embeddings to domains with different language usage from the training domain.

Our research is motivated by the idea that utilising existing language knowledge that is invariant across domains and language uses may aid the learning of more universal (i.e.\ domain independent) word embeddings. Incorporating background knowledge into statistical learning to improve model robustness or transferability is a topic of great interest in NLP \citep{Bian2014, Faruqui2015, Fried2015a}. However, there is no clear way to deal with the ensuing trade-off between preserving the distributional information and incorporating external knowledge. We explore whether and how this trade-off can be mitigated through a novel approach that respects the distributional learning regime while adding examples that are reflective of a particular property found in the background knowledge.

This paper has two main contributions. First, a novel approach to synthetic text data augmentation to inject linguistic knowledge into word embeddings, which leverages semantic information from pre-existing human constructed knowledge bases. Second, a systematic evaluation of the effect that this procedure has on the learned word embeddings.

In section \ref{sec:related} we briefly present some of the work that has been done in terms of incorporating background knowledge into NLP models. Section \ref{sec:skipgram} gives a technical description of our experimental setup and in particular our process of data augmentation. Section \ref{sec:evaluation} evaluates the trained word embeddings from two perspectives: extrinsic, which measures the performance of the embeddings on downstream tasks, specifically a document classification task; and intrinsic, which focuses on the properties of the learned embeddings. Section \ref{sec:discussion} concludes with a discussion about our findings.

\section{Related Work}
\label{sec:related}

The increasing adoption of word embeddings, particularly following the Word2Vec models \citep{Mikolov2013b, Mikolov2013}, has been followed by a large body of research attempting to understand the properties that are learned by these embedding models, their limitations, and possible ways to extend them. Our work lies at the intersection of two research avenues in NLP: knowledge injection into word embeddings, and textual data augmentation.

\subsection{Knowledge Injection}

We use the term \textit{knowledge injection} to refer to the incorporation of information from external human-constructed knowledge bases into a statistical learning process. The integration of knowledge and distributional learning in the context of word embeddings has been attempted with two general approaches:

\paragraph{Post-processing} Post-processing starts with a set of pre-trained word embeddings, which it subsequently transforms (or \textit{fine-tunes}) to reflect specific relationships contained in a knowledge base or ontology. These transformations, however, are commonly constrained to prevent them from washing away the distributional information from the original embeddings. \textit{Retrofitting} \citep{Faruqui2015} brings pre-trained vectors closer together in embedding space if their corresponding words appear in a relationship in the knowledge base, while minimising the distance they move from their original position. Similar work by \citet{Mrksic2016} and \citet{Mrksic2017} imposes linguistic constraints of synonym attraction and antonym repel, together with a vector space preservation term, to learn a new word embedding matrix from the existing one. More recent work by \citet{Vulic2018a} applies linguistic constraints to learn \textit{hyponymy-hypernymy} relationships, while \citet{Vulic2018} extends these ideas to \textit{unseen} words (i.e. not appearing in the knowledge base) by incorporating a neural network that learns a mapping function that applies the transformations of the ``seen subspace'' to the ``unseen subspace''.

\paragraph{Objective Function Modification} This second approach consists of incorporating an additional (\textit{regularisation}) term to an existing training objective, where the added term enforces specific semantic relationships gathered from knowledge bases. The resulting loss functions takes the general form:

\[
    \mathcal{L} = \mathcal{L}_{distributional} + \lambda \mathcal{L}_{knowledge}
\]

\noindent where \( \mathcal{L} \) is the total loss term, \( \mathcal{L}_{distributional} \) is a distributional or data-driven loss term, and \( \mathcal{L}_{knowledge} \) is a loss term depending on the incorporation of external knowledge, weighted by \( \lambda \).

\citet{Yu} add a linear term to the two Word2Vec training objectives, Skip-gram a nd Continuous Bag-of-Words (CBOW) \citep{Mikolov2013b}, which aim to predict one word from a related word, where this relationship is extracted from a semantic knowledge base. The work by \citet{Bian2014} is based on the CBOW model from Word2Vec (where the target word is predicted from a context vector) and it incorporates auxiliary loss terms which correspond to specific relationships obtained from different knowledge bases, such that given the context of a focus word, the model needs to learn to predict the focus word, as well as its synonyms, Part-of-Speech (POS) tags, categories, hyponyms, etc. To incorporate \textit{relational} information into distributional learning of word embeddings, \citet{Fried2015a} propose a loss function that combines a distributional (\textit{neural language model}) loss term with a graph distance loss, which is based on a graph constructed from \textit{synset} relationships in the \textit{WordNet} \citep{GeorgeA.Miller1995} knowledge base. Similarly, \citet{Liu2015a} add a term based on semantic similarity ranking for word triplets, defined in terms of their corresponding embeddings. \citet{Jiang2018} propose a \textit{hybrid} loss function that combines a distributional loss term, such as any of the Word2Vec objective functions, with a \textit{knowledge enriched} loss based on learning the relationships in a ``word reading difficulty'' graph.

\subsection{Data Augmentation}

The term ``data augmentation'' was initially used in computer vision and is currently a widely used technique in that field \citep{Ciresan2010, Krizhevsky2012, Wong2016}. In general terms, it refers to the application of some label-preserving transformation to the data (e.g. rotations \citep{Rowley1998}, image scaling, or \textit{elastic distortions} \citep{Simard2003} which are parameterised deformations of an image to emulate oscillations of hand muscles in handwritten digits). These transformations seek to increase the size and diversity of the training data while preserving its semantic content. The aim behind this technique in computer vision is to improve the robustness of the model and to prevent overfitting \citep{Simard2003}. However, the concept of a ``label-preserving transformation'' which can be described as a parametric function is not clearly defined for text data, and is furthermore hampered by polysemy, context dependence, and other common features inherent to language. Hence, text data augmentation for most NLP tasks remains an open research question.

\citet{Banko2001} applied a simple augmentation procedure to the task of choosing the correct word from a set of \textit{confusable} words such as \textit{your} and \textit{you're}. This procedure consisted of replacing all occurrences of confusable words in large unannotated text corpora with a marker which the learning algorithm must fill in with the correct member of the confusion set. With this simple augmentation procedure they showed nearly two decades ago that the size and diversity of a training dataset can have a large impact on the performance of the model, regardless of the training algorithm that is used.

In the context of semi-supervised learning, \citet{Xie2019} propose two methods for \textit{advanced data augmentation} (a opposed to \textit{simple} data augmentation methods which inject noise to unlabelled examples) for text classification: back-translation, a technique developed by \citet{Sennrich2016} which produces paraphrases by automatically translating an input text into a \textit{pivot} language and subsequently translates it back to the original language; and word replacement with TF-IDF, which replaces uninformative words (low TF-IDF) with a randomly sampled word, while preserving informative ones (high TF-IDF).

The specific strand of data augmentation under which our work falls is what we call \textit{knowledge-driven} data augmentation. In NLP this most commonly consists of a series of lookup operations into a lexical knowledge base, such as a thesaurus, to replace words or phrases with semantically equivalent ones. The idea of data augmentation through synonym replacement is not new and has appeared in the work of \citet{Zhang2016b} and \citet{Coulombe2018} to increase the size of the training data for supervised text classification tasks. Also within the context of data augmentation for text classification, \citet{Kobayashi2018} describes a \textit{contextual augmentation} technique which performs synonym or related word replacement based on label-conditioned language models. 

Our data augmentation approach does not depend on any additional training or labels, it performs synonym augmentation solely in terms of word frequencies within a corpus. To the best of our knowledge, our work is the first to look at the effects of knowledge-driven data augmentation on the unsupervised learning of word embeddings.

\section{Knowledge-Augmented Skip-gram}
\label{sec:skipgram}

Our work builds upon the Word2Vec Skip-gram + Negative sampling model described in \citep{Mikolov2013} which, in short, attempts to predict \textit{context} words (i.e.\ words appearing within the context window of a word) given a \textit{focus} word, \( p(w_{context} \bigm| w_{focus}) \). Negative sampling approximates the full softmax calculation of this probability, which speeds up computation by distinguishing between the context word and a noise signal.

Word2Vec is based on the \textit{distributional hypothesis} \citep{Harris1954}, which is the assumption that a word's meaning can be inferred from its context (i.e.\ accompanying words). Following this hypothesis our idea is that, given that synonyms by definition have similar meanings, this semantic similarity can be enforced by placing synonyms in the same contexts. This is achieved by adding examples of synonyms appearing in the same context as the original words and, by doing so, enforcing vector representations of synonyms to be closer together in embedding space.

In order to test the effect of incorporating external knowledge into the training dataset we recreated the original Skip-gram + Negative sampling experimental setup as closely as possible. We worked with a subset (160 randomly sampled books) of the Gutenberg Project dataset, a repository of classic books ranging from literature to science\footnote{Note that the language from this dataset will differ significantly from that used in the original Word2Vec given the difference in publication times and medium (i.e. published books vs. news articles).}.

\begin{figure*}[t]
\centering
\includegraphics[width=.8\linewidth]{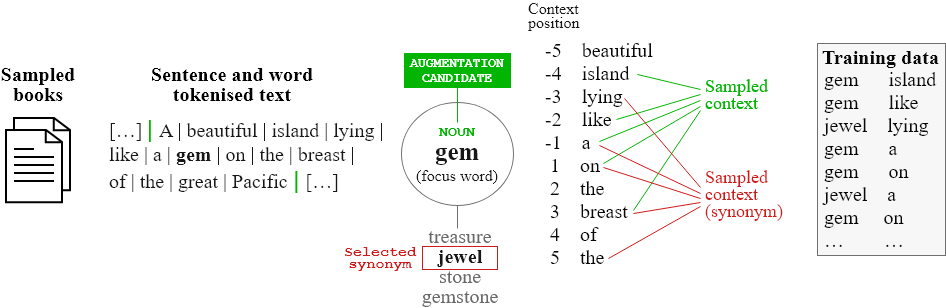}
\caption{Synonym augmentation pipeline}
\label{fig:pipeline}
\end{figure*}

The full setup, illustrated in figure \ref{fig:pipeline}, is as follows. First we tokenised the text corpus into sentences, so that no word pair spanned more than one sentence\footnote{This step, which isn't described in the literature, follows the assumption that sentences are, in Gottlob Frege's words,  the ``proper means of expression for a thought'' \citep{May2006}, so \textit{semantically valid} contexts should not span more than one sentence.}. Then we constructed the word pair dataset following the procedure of the Skip-gram model, where every word in the training corpus is treated as a focus word. Finally, for each of its context words we constructed a word pair, which constituted a training datapoint.

To decrease the relative frequency of words that are further away from the focus word, the original model randomly samples a context window size given an initial \textit{maximum} context size \( C \). We performed an equivalent sampling procedure by constructing all word pairs (with context size \( C \)) and sampled each word pair based on its context position \( c \) with probability \( p(c) = \frac{C - c + 1}{C} \).

After all of this was done, we performed our knowledge integration extension. First, we selected \textit{candidate words} by going through every focus word and selecting only adjectives, adverbs, nouns or verbs. Next, we queried WordNet, a lexical knowledge base, for synonyms of the candidate words. Since some words have several synonyms, we sampled the \textit{augmented data} in order to get a single synonym per candidate word. This sampling was based on word frequencies, such that more frequent words in the training corpus were more commonly selected as synonyms. For example (see figure \ref{fig:pipeline}), after selecting \textit{gem} as a candidate word and querying for its synonyms, the probability of selecting \textit{jewel} as the augmenting synonym would be proportional to the frequency with which the term \textit{jewel} appears in the training data.

Finally, as is common practice in NLP research, we constructed a subset of the full vocabulary of 91,829 distinct words from our training data. This is done to reduce the number of distinct words, as well as remove very rare words and misspellings from our training data. Our vocabulary reduction strategy was based on setting a frequency threshold, such that any word that appeared fewer than \( n \) times got pruned from the vocabulary. We performed our experiments at three vocabulary sizes, which we call after their cutoff points \textit{vocabulary-3} (45,069 words), \textit{vocabulary-7} (30,369 words), and \textit{vocabulary-20} (17,776 words). The original Word2Vec model uses a frequency threshold of \( n=5 \) and a vocabulary size of 3,000,000.

The experimental setup otherwise followed the description in \citet{Mikolov2013} and \citet{Mikolov2013b}. We used an embedding size of 300, context size  \( C=5 \), \( k=5 \) negative samples (which the paper suggests for smaller datasets), trained for 10 epochs with a learning rate of 0.01 since that resulted in a convergent training phase\footnote{We could not find these details for the original work, so we used common choices and validated that the learning converged without overfitting}, and a batch size of 10 to speed up training. We skipped the \textit{subsampling} phase described in \citep{Mikolov2013b} because our dataset was not large enough to warrant an additional speed up, and because our data augmentation strategy already down-weights stop words by leaving them out of the augmentation process. We trained our word embeddings with a fixed \textit{augmentation ratio} of \( r=0.25 \) (i.e. the training data consisted of 75\% \textit{natural} data and 25\% synonym-augmented data), which we empirically found to be the best performing ratio.

In order to ensure a systematic evaluation we trained models for each vocabulary with two different initialisations: randomly initialised, and Word2Vec initialised\footnote{Word2Vec is a bilinear model, i.e. it consists of two distinct \textit{embedding units} referred to as \textit{input} and \textit{output} embeddings. However, we were only able to find the pre-trained input embedding unit, so we initialised the output embeddings randomly.}. Furthermore, to accurately gauge the effects of the augmentation process we trained a version of all of these models without any data augmentation.

\section{Evaluation}
\label{sec:evaluation}

Despite the mainstream adoption of word embeddings across (and beyond) NLP tasks, there is still no single agreed upon method to measure their representational quality. Rather, there are a suite of different metrics to determine the usefulness of word embeddings, or the degree to which they reflect specific syntactic or semantic phenomena. These metrics can be divided into two main categories, as proposed by \citet{Schnabel2015}: \textit{extrinsic evaluation}, evaluating word embeddings based on their effectiveness as representations for downstream tasks; and \textit{intrinsic evaluation}, focusing on how the internal geometry of the word embeddings reflects specific linguistic relationships.

\subsection{Extrinsic Evaluation}

Word embeddings can be extrinsically evaluated by using them as input representations for NLP tasks such as sentiment analysis, part-of-speech tagging, named entity recognition, recognising textual entailment, among many others.

We focus on the Word Mover's Distance (WMD) \citep{Kusner2015}, document classification task for the extrinsic evaluation because of its high dependence on word embeddings. WMD is a distance metric between documents calculated by solving a transport problem based on the \textit{Earth Mover's Distance} \citep{Rubner1998}, in which the word distribution of the first document has to be transformed into the word distribution of the second, where the \textit{travel cost} between two words is the Euclidean distance between their corresponding embeddings. The document classification task is then solved with K-Nearest Neighbours classification based on the WMD between document pairs.

The original WMD paper reports accuracies on different datasets. We focused on the 20 News Groups dataset \citep{Joachims1996}, which comprises 18,846 news articles from 20 relatively well-balanced classes (or \textit{news groups}). In their dataset description they report evaluating on 11,293 articles from the dataset. This seems to correspond to the training subset as obtained from the SciKit Learn \citep{Pedregosa2011} implementation\footnote{\url{https://scikit-learn.org/stable/modules/generated/sklearn.datasets.fetch\_20newsgroups.html}}, which contains 11,314 articles. The paper mentions only focusing on the most common 500 words in each document as a speed up technique. We tested our embeddings on the full article to minimise the impact of a change in our vocabulary size.

Because of computation times, we skipped the search for the optimal \( K \) described by \citet{Kusner2015}; we instead kept a constant value of \( K=10 \) to compare all of our different learned embeddings. The accuracies for the KNN+WMD document classification on the 20 News Groups dataset are reported in table \ref{tab:wmd_accuracies}.

\begin{table*}[]
\centering
\resizebox{\linewidth}{!}{%
\begin{tabular}{l|c|cc|cc}
         & Word2Vec & \multicolumn{2}{c|}{Word2Vec init.} & \multicolumn{2}{c}{Random init.} \\ 
         & (base)        & No Syns.              & Syns.            & No Syns.        & Syns.    \\ \hline
Voc-3    & .607 (\( \pm \).0090)    & .654 (\( \pm \).0088)    & .651 (\( \pm \).0088)    & .583 (\( \pm \).0091)   & .577 (\( \pm \).0091) \\
Voc-7    & .611 (\( \pm \).0090)   & .649 (\( \pm \).0088)         & .649 (\( \pm \).0088)     & .575 (\( \pm \).0091)   & .574 (\( \pm \).0091) \\
Voc-20   & .597 (\( \pm \).0090)   & .643 (\( \pm \).0088)         & .640 (\( \pm \).0088)     & .570 (\( \pm \).0091)   & .558 (\( \pm \).0092) \\ \hline
Voc-W2V  & .783 (\( \pm \).0076)   & -                  & -              & -            & -
\end{tabular}
}
\caption{KNN+WMD accuracies for the 20 News Groups dataset (intervals for a confidence level of 95\% are given in parenthesis). \textit{W2V (base)} is the unaltered Word2Vec model pruned to different vocabulary sizes. \textit{W2V init.} and \textit{Rand. init.} refer to the different initialisation points, and are divided into non-augmented (\textit{NoSyn}) and augmented (\textit{Syn}) training data. \textit{Voc-W2V} is the result from the pre-trained Word2Vec embeddings with the full 3,000,000 word vocabulary, which in the original WMD paper by \citet{Kusner2015} is reported to have an accuracy of .73.}
\label{tab:wmd_accuracies}
\end{table*}

The first thing to note in the results from table \ref{tab:wmd_accuracies} is the effect of the vocabulary size on the performance of the model on the KNN+WMD task. The first column, which corresponds to the unaltered Word2Vec embeddings, shows a large bump in classification accuracy when using a larger subset of the same word embeddings (3M words as opposed to 45K/30K/17K words). Given that the only difference between the rows in the first column is the size of the vocabulary, while the embeddings remain identical, this suggests that the information contained in rare words is highly relevant to the document classification task.

On our vocabularies we observe a clear gain in classification accuracy when using the word embeddings we initialise with the Word2Vec embeddings and subsequently train on the Gutenberg data compared to the unaltered Word2Vec embeddings. Our randomly initialised embeddings, on the other hand, achieve a lower accuracy than both the unaltered Word2Vec and our Word2Vec initialised embeddings. The accuracies obtained using our knowledge-augmented embeddings, while slightly lower than those obtained by the non-augmented embeddings, do not differ significantly.

\subsection{Intrinsic Evaluation}

We performed two separate intrinsic evaluations on our learned word embeddings. For the first evaluation we calculated the correlation between a human-assigned similarity score for a pair of words and the distance between their corresponding embeddings. The second evaluation compared embedding distance distributions for three sets of word pairs.

\subsubsection{Similarity-Distance Correlation}

For the similarity-distance correlation we used two popular datasets: WordSim353 \citep{Finkelstein2002}, which as its name implies consists of 353 word pairs with a corresponding similarity score. This dataset has been divided into two subsets to distinguish between \textit{similarity} (a measure of how alike two words are) and \textit{relatedness} (how closely related, but not necessarily similar, two words are); and SimLex999 \citep{Hill2015}, a more recent dataset containing 999 word pairs with human annotated similarity scores. Unlike WordSim353, this dataset focuses exclusively on similarity, so the scores ignore any \textit{relatedness} or \textit{association} between the words. The example provided in SimLex999 makes a distinction between \textit{coast} and \textit{shoreline}, which it identifies as being similar, and \textit{clothes} and \textit{closet}, which are related, but not similar.

\begin{figure}[t]
	\centering
	\includegraphics[width=\linewidth]{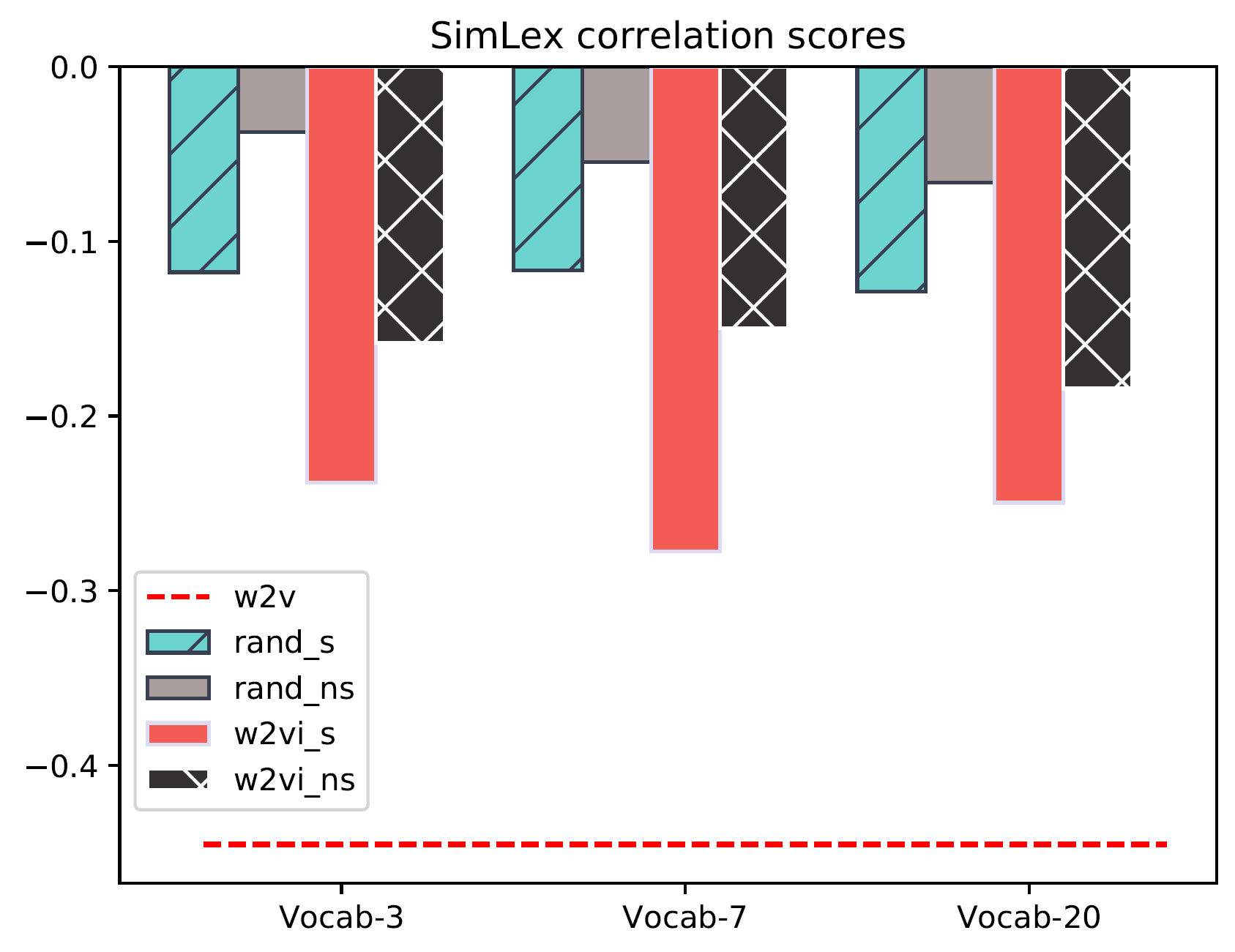}
    \caption{Spearman's \( \rho \) between embedding distances and word similarity scores on the SimLex999 dataset on our three vocabulary sizes with a constant data augmentation ratio of 25\%. \textit{w2v} shows the performance of the unaltered Word2Vec embeddings.}
	\label{fig:correl_vocabs}
\end{figure}

We report Spearman's rank-correlation coefficient, or Spearman's \( \rho \), between word similarity scores from SimLex999 and WordSim353, and the cosine distances between the corresponding word embeddings\footnote{We omit the correlation results for Euclidean distance, since they consistently give smaller (magnitude) correlation scores than cosine distance for all models and all metrics.}. To allow a more direct comparison, we only use the word pairs that appear in the smallest vocabulary (899/999 for SimLex999, 184/251 for WordSim353-rel, and 142/202 for WordSim353-sim) so all results are reported on the same set of word pairs (this also explains why the baseline Word2Vec results are exactly the same on all three vocabulary sizes as shown in figure \ref{fig:correl_vocabs}).

\begin{figure}[t]
	\centering
	\includegraphics[width=\linewidth]{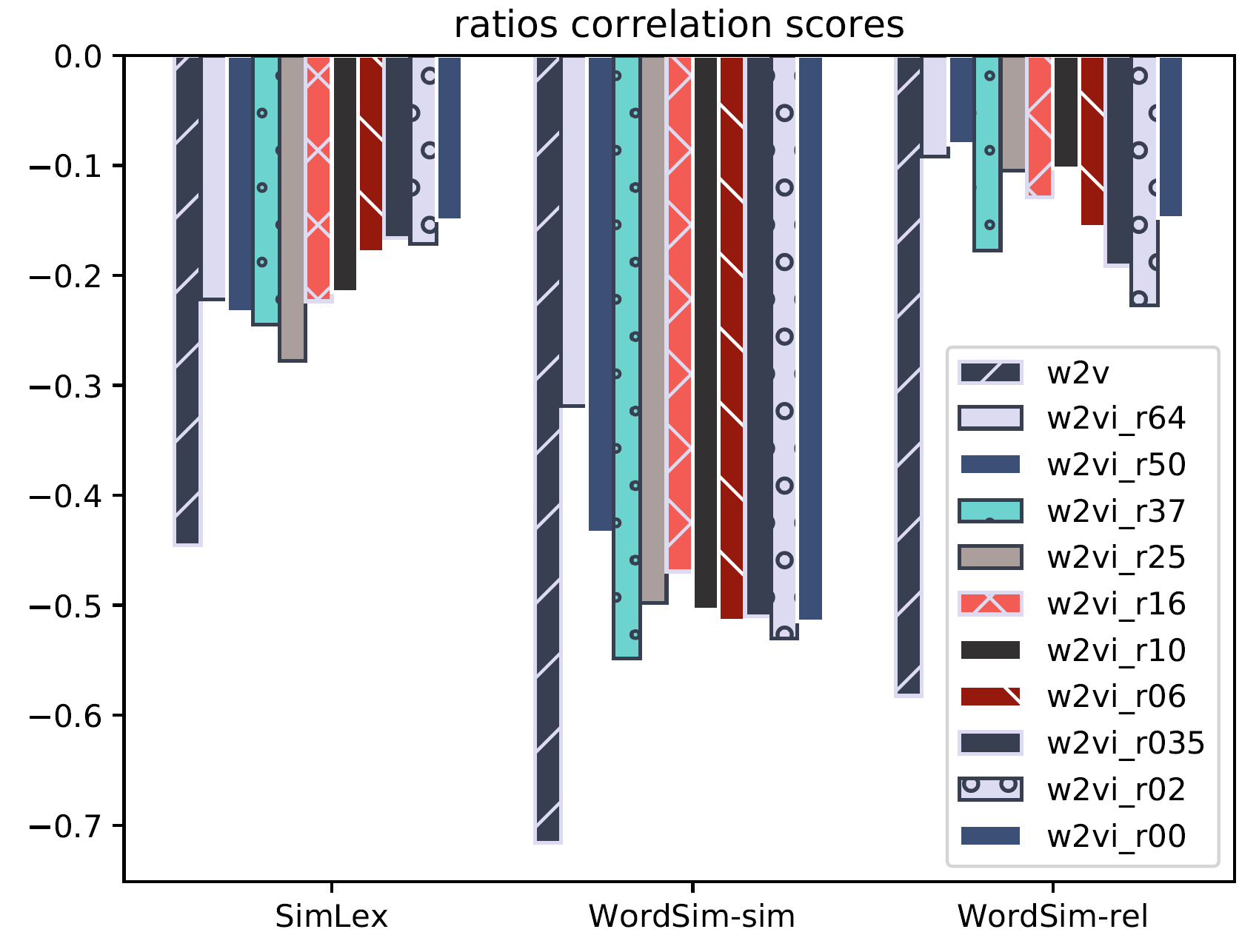}
	\caption{Spearman's \( \rho \) for varying ratios of data augmentation (0\%, 2\%, 3.5\%, 6\%, 10\%, 16\%, 25\%, 37\%, 50\%, 64\%) and for the unaltered Word2Vec embeddings (\textit{w2v}), between embedding distances and word similarity or relatedness scores for three datasets: SimLex999, WordSim353-similarity, and WordSim353-relatedness.}
	\label{fig:correl_ratios}
\end{figure}

In figure \ref{fig:correl_ratios} we show the correlation scores for Word2Vec-initialised vocabulary-7 embeddings with different augmentation ratios. The unaltered Word2Vec embeddings achieve the best correlation scores (lower is better) across all three similarity-distance correlation tasks. This could again be due to the size of the training data used for the original Word2Vec model, which is likely to contain many more samples of the words in these datasets. The drop in correlation on the non-augmented version of our embeddings (\textit{w2vi\_r00}) indicates that our subsequent training on the Gutenberg data is washing away some of the information contained in the original embeddings. The relatedness dataset, \textit{WordSim-rel}, experiences the largest drop in correlation score. This could also be related to the size of the training data, since learning relatedness relationships requires more contextual examples. Sentence tokenisation could also be hurting this measure, removing inter-sentence information and thus reducing the context window size (which is limited by the span of the focus word's sentence).

Regarding the effect of data augmentation, we observe that WordSim-sim obtains the highest correlation scores for all embeddings and seems to be the least affected by the augmented data, until the augmentation ratio is increased to 50\% and 64\%, at which point the results experience a significant drop. WordSim-rel seems to benefit from a small amount of augmentation, 2\%, but there is no clear trend since results vary for different ratios of augmentation. SimLex999 shows a consistent improvement in correlation scores as the augmentation ratio increases until it reaches a peak at 25\%, after which results start to gradually decline as the ratio increases. For all three datasets, there appears to be a degree of data augmentation that improves upon the correlation scores of the non-augmented model (\textit{w2vi\_r00}).

We analyse correlation scores for our three different vocabulary sizes in figure \ref{fig:correl_vocabs}. We focus on SimLex999 since it is a more recent dataset which focuses entirely on similarity, is larger in size, has high coverage with our smallest vocabulary (\( \thicksim \)90\%), and is a harder task overall. Figure \ref{fig:correl_vocabs} shows a consistent improvement in correlation scores after augmentation for both randomly initialised and Word2Vec initialised models across all three vocabulary sizes.

\subsubsection{Embedding Space Analysis}

To gain some insight into the underlying geometric properties of the learned embedding spaces, we analyse the distance distributions with respect to three sets of word pairs:
\begin{itemize}
    \item \textbf{Synonyms} - a set of distinct synonym pairs from our augmented data
    \item \textbf{Contextual} - a set of word pairs appearing within each other's context in the \textit{natural} data
    \item \textbf{Random} - a set of randomly sampled word pairs from our vocabulary
\end{itemize}

\begin{figure*}[t]
	\centering
	\begin{subfigure}{.48\linewidth}
		\centering
		\includegraphics[width=\linewidth]{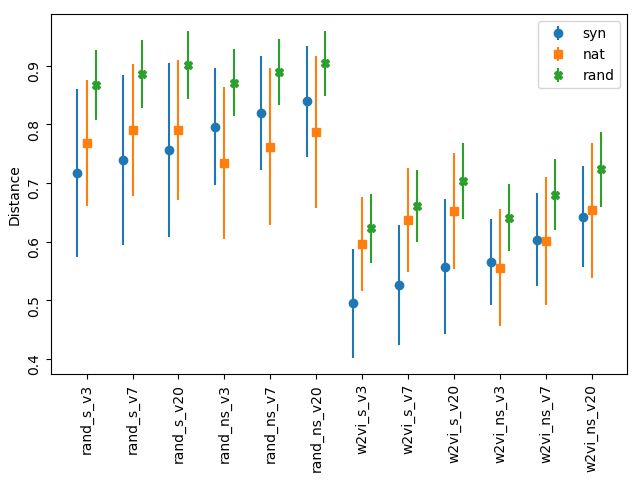}
        \caption{Model variations\label{fig:all_dists}}
	\end{subfigure}
	\begin{subfigure}{.48\linewidth}
		\centering
		\includegraphics[width=\linewidth]{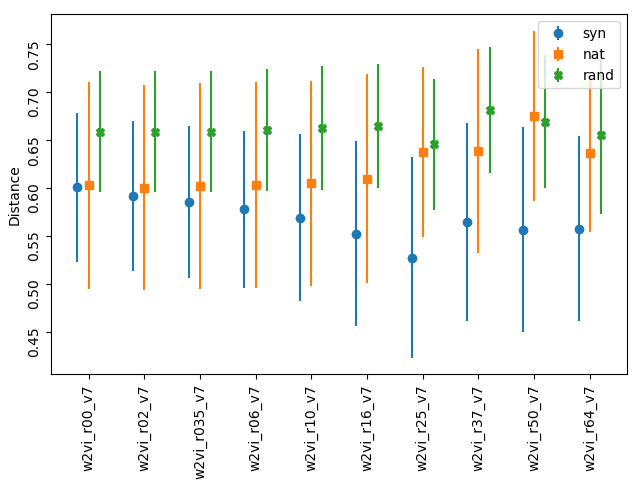}
        \caption{Augmentation ratio variations\label{fig:ratios_dists}}
	\end{subfigure}
    \caption{Word embedding pairwise cosine distance mean and standard deviation for three sets of word pairs: synonyms (\textit{syn}), contextual (\textit{nat}), and random (\textit{rand}). As in our other plots, \textit{rand\_} and \textit{w2vi\_} refer to the initialisation, \textit{s\_} and \textit{ns\_} correspond to augmented and non-augmented respectively, \textit{v\_} indicates the vocabulary size, and \textit{r\_} the augmentation ratio.}
	\label{fig:mean_var_dists}
\end{figure*}

Figure \ref{fig:mean_var_dists} shows the mean and standard deviation of the cosine distance distributions on each set of word pairs for model variations (figure \ref{fig:all_dists}) and for different augmentation ratios (figure \ref{fig:ratios_dists}). In figure \ref{fig:all_dists} we confirm that our data augmentation approach is indeed bringing pairs of synonyms closer together than \textit{contextual} word pairs (i.e. words that appear within the same context in the text corpus), while distances between random word pairs remain mostly unaffected (compare \textit{s\_} against \textit{ns\_} columns in figure \ref{fig:all_dists}). This holds for all vocabulary sizes and initialisations. This is a positive observation since ideally we should see the synonym distances being the smallest, contextual distances in the middle, and random distances being the largest. Notice that distances for all sets of word pairs are substantially smaller for the Word2Vec initialised models than for the randomly initialised ones.

In figure \ref{fig:ratios_dists}, when going from no augmentation up to a ratio of 16\%, the synonym pairs are getting gradually closer while the contextual and random pair distances are mostly unaffected. Synonym distances become smallest at a ratio of 25\%, but this comes at the expense of random words getting closer and contextual words moving further apart. Augmentation ratios above 25\% increase the synonym distances up to a roughly constant distribution, and also affect the contextual distances, which become largest at 50\%, and the random distances, which gradually decrease from 37\% onwards.

\section{Discussion}
\label{sec:discussion}

In both the KNN+WMD text classification task and the similarity-distance correlation, the full vocabulary Word2Vec model (\textit{Voc-W2V}) outperforms all our models, which was expected due to the differences in the size of the vocabulary and the size of the training data. Larger vocabularies result in better coverage of the language in the target task, which in the case of KNN+WMD implies that rarer words might carry more relevant information for the document classification task. 

A significant difference between the original Word2Vec setup and ours is the size of the training data: we train on a subset of the Gutenberg dataset, which is orders of magnitude smaller than the Google News dataset. Our results show Word2Vec initialised embeddings outperform the randomly initialised ones in both the classification and correlation tasks, which supports the claim that the size of the training data is one of the main indicators of the quality of the learned embeddings \citep{Mikolov2013}. Using a smaller training dataset implies that our model trained from scratch does not capture the less frequent uses of words, so the language model we learn from this data is not very diverse and ignores rare contexts. We therefore hypothesise that training our models on larger training datasets would further improve their quality.

Our Word2Vec initialised embeddings significantly outperform the cropped-vocabulary Word2Vec embeddings in the KNN+WMD task (see table \ref{tab:wmd_accuracies}, columns 1 to 3). During training, the Word2Vec model optimises the informational capacity of the embedding space, so we hypothesise that removing vocabulary items from the fully trained embedding space creates ``holes'' which make the classification task harder since the word distributions need to find close embeddings to move the probability mass to. This suggests that subsequent training of the cropped-vocabulary Word2Vec embeddings ``fills in the holes'' in the embedding space by adapting it to the new vocabulary size. Note that performance improves after subsequent training despite the linguistic domain of the Gutenberg dataset being further away from 20 News Groups than the original training data for Word2Vec, which is Google News\footnote{The tasks we report here are based on contemporary language usage and the vast majority of books we sampled from the Gutenberg dataset are over 100 years old and favour more specialised and elaborate language.}.

The combination of evaluation metrics that we use shows that improving word embeddings for one task does not necessarily improve their overall quality, and might affect their results for other tasks. Our augmentation approach improves the intrinsic evaluation of the embeddings while not significantly impacting the performance on the text classification task. The ``ideal'' augmentation ratio, however, seems to vary from task to task, as seen in figure \ref{fig:mean_var_dists}, where certain augmentation ratios produce more \textit{semantically desirable} embeddings, where synonyms are closer together than contextual pairs, and the largest distances occur withing randomly sampled word pairs.

To summarise, in this paper we propose an approach to knowledge injection into word embeddings that respects the distributional characteristics of the original Skip-gram language model. Through a systematic comparison between training regimes with and without data augmentation and with varying augmentation ratios, we show that this approach improves the \textit{intrinsic} characteristics of the learned embeddings while their performance in a text classification task is not significantly altered. This approach requires additional pre-processing, but it has no impact on model complexity or training times. Unlike post-processing techniques, our approach is applied during training of the word embeddings, so to investigate its full potential and limitations it needs to be applied to a larger scale training dataset. In future work we plan to explore this augmentation technique on the full dataset used for the original Word2Vec model. We also plan to explore the applicability of this approach to relatedness relationships between words.

\section*{Acknowledgments}
The authors acknowledge the use of the UCL Myriad High Throughput Computing Facility (Myriad@UCL), and associated support services, in the completion of this work.
Diego Ramírez-Echavarría would like to express his gratitude to the Mexican National Council for Science and Technology (CONACYT) for funding his PhD project.

\bibliography{library}

\begin{thebibliography}{34}
\expandafter\ifx\csname natexlab\endcsname\relax\def\natexlab#1{#1}\fi

\bibitem[{Banko and Brill(2001)}]{Banko2001}
Michele Banko and Eric Brill. 2001.
\newblock \href {https://www.aclweb.org/anthology/H01-1052} {{Mitigating the
  Paucity-of-Data Problem: Exploring the Effect of Training Corpus Size on
  Classifier Performance for Natural Language Processing}}.
\newblock In \emph{Proceedings of the First International Conference on Human
  Language Technology Research}. ACL.

\bibitem[{Bian et~al.(2014)Bian, Gao, and Liu}]{Bian2014}
Jiang Bian, Bin Gao, and Tie-Yan Liu. 2014.
\newblock \href
  {https://link.springer.com/content/pdf/10.1007{\%}2F978-3-662-44848-9{\_}9.pdf}
  {{Knowledge-Powered Deep Learning for Word Embedding}}.
\newblock \emph{Machine Learning and Knowledge Discovery in Databases}, pages
  132--148.

\bibitem[{Ciresan et~al.(2010)Ciresan, Meier, Gambardella, and
  Schmidhuber}]{Ciresan2010}
Dan~Claudiu Ciresan, Ueli Meier, Luca~Maria Gambardella, and J{\"{u}}rgen
  Schmidhuber. 2010.
\newblock \href {http://arxiv.org/abs/1003.0358v1} {{Deep Big Simple Neural
  Nets Excel on Hand-written Digit Recognition}}.
\newblock \emph{Neural computation}, 22(12):3207--3220.

\bibitem[{Coulombe(2018)}]{Coulombe2018}
Claude Coulombe. 2018.
\newblock \href {https://arxiv.org/abs/1812.04718} {{Text Data Augmentation
  Made Simple By Leveraging NLP Cloud APIs}}.
\newblock \emph{arXiv}.

\bibitem[{Faruqui et~al.(2015)Faruqui, Dodge, Jauhar, Dyer, Hovy, and
  Smith}]{Faruqui2015}
Manaal Faruqui, Jesse Dodge, Sujay~K Jauhar, Chris Dyer, Eduard Hovy, and
  Noah~A Smith. 2015.
\newblock \href
  {https://www.cs.cmu.edu/{~}hovy/papers/15HLT-retrofitting-word-vectors.pdf}
  {{Retrofitting Word Vectors to Semantic Lexicons}}.
\newblock In \emph{Proceedings of NAACL}.

\bibitem[{Finkelstein et~al.(2002)Finkelstein, Gabrilovich, Matias, Rivlin,
  Solan, Wolfman, and Ruppin}]{Finkelstein2002}
Lev Finkelstein, Evgeniy Gabrilovich, Yossi Matias, Ehud Rivlin, Zach Solan,
  Gadi Wolfman, and Eytan Ruppin. 2002.
\newblock \href {http://www.cs.tau.ac.il/{~}ruppin/p116-finkelstein.pdf}
  {{Placing Search in Context: The Concept Revisited}}.
\newblock \emph{ACM Transactions on information systems}, 20(1):116--131.

\bibitem[{Fried and Duh(2015)}]{Fried2015a}
Daniel Fried and Kevin Duh. 2015.
\newblock \href {http://arxiv.org/abs/1412.4369v3} {{Incorporating Both
  Distributinoal and Relational Semantics in Word Representations}}.
\newblock \emph{arXiv}.

\bibitem[{{George A. Miller}(1995)}]{GeorgeA.Miller1995}
{George A. Miller}. 1995.
\newblock \href {https://doi.org/10.1145/219717.219748} {{WordNet: A Lexical
  Database for English}}.
\newblock \emph{Communications of the ACM}, 38(11):39--41.

\bibitem[{Harris(1954)}]{Harris1954}
Zellig~S Harris. 1954.
\newblock \href {https://doi.org/10.1080/00437956.1954.11659520}
  {{Distributional Structure}}.
\newblock \emph{WORD}, 10(3):146--162.

\bibitem[{Hill et~al.(2015)Hill, Reichart, and Korhonen}]{Hill2015}
Felix Hill, Roi Reichart, and Anna Korhonen. 2015.
\newblock \href {https://doi.org/10.1162/COLI} {{SimLex-999: Evaluating
  Semantic Models With (Genuine) Similarity Estimation}}.
\newblock \emph{Computational Linguistics}, 41(4):665--695.

\bibitem[{Jiang et~al.(2018)Jiang, Gu, Yin, and Chen}]{Jiang2018}
Zhiwei Jiang, Qing Gu, Yafeng Yin, and Daoxu Chen. 2018.
\newblock \href {https://aclweb.org/anthology/C18-1031} {{Enriching Word
  Embeddings with Domain Knowledge for Readability Assessment}}.
\newblock In \emph{Proceedings of the 27th International Conference on
  Computational Linguistics}, pages 366--378, Santa Fe, NM, USA. ACL.

\bibitem[{Joachims(1997)}]{Joachims1996}
Thorsten Joachims. 1997.
\newblock \href {https://dl.acm.org/citation.cfm?id=657278} {{A Probabilistic
  Analysis of the Rocchio Algorithm with TFIDF for Text Categorization}}.
\newblock In \emph{Proceedings of the Fourteenth International Conference on
  Machine Learning}, pages 143--151, Pittsburgh, PA, USA. Carnegie Mellon
  University.

\bibitem[{Kobayashi(2018)}]{Kobayashi2018}
Sosuke Kobayashi. 2018.
\newblock \href {https://www.aclweb.org/anthology/N18-2072/} {{Contextual
  Augmentation: Data Augmentation by Words with Paradigmatic Relations}}.
\newblock In \emph{Proceedings of NAACL-HLT 2018}, pages 452--457. ACL.

\bibitem[{Krizhevsky et~al.(2012)Krizhevsky, Sutskever, and
  Hinton}]{Krizhevsky2012}
Alex Krizhevsky, Ilya Sutskever, and Geoffrey~E Hinton. 2012.
\newblock \href
  {http://papers.nips.cc/paper/4824-imagenet-classification-with-deep-convolutional-neural-networ}
  {{ImageNet Classification with Deep Convolutional Neural Networks}}.
\newblock In \emph{Advances in neural information processing systems}, pages
  1097--1105. NeurIPS.

\bibitem[{Kusner et~al.(2015)Kusner, Sun, Kolkin, and Weinberger}]{Kusner2015}
Matt~J Kusner, Yu~Sun, Nicholas~I Kolkin, and Kilian~Q Weinberger. 2015.
\newblock \href {http://proceedings.mlr.press/v37/kusnerb15.pdf} {{From Word
  Embeddings To Document Distances}}.
\newblock In \emph{International Conference on Machine Learning}, pages
  957--966.

\bibitem[{Liu et~al.(2015)Liu, Jiang, Wei, Ling, and Hu}]{Liu2015a}
Quan Liu, Hui Jiang, Si~Wei, Zhen-Hua Ling, and Yu~Hu. 2015.
\newblock \href {https://www.aclweb.org/anthology/P15-1145/} {{Learning
  Semantic Word Embeddings based on Ordinal Knowledge Constraints}}.
\newblock In \emph{Proceedings of the 53rd Annual Meeting of the Association
  for Computational Linguistics and the 7th International Joint Conference on
  Natural Language Processing}, pages 1501--1511. ACL.

\bibitem[{May(2006)}]{May2006}
Robert May. 2006.
\newblock \href
  {https://www.researchgate.net/publication/261824012{\_}The{\_}Invariance{\_}of{\_}Sense}
  {{The Invariance of Sense}}.
\newblock \emph{The Journal of philosophy}, 103(3):111--144.

\bibitem[{Mikolov et~al.(2013{\natexlab{a}})Mikolov, Chen, Corrado, and
  Dean}]{Mikolov2013b}
Tomas Mikolov, Kai Chen, Greg Corrado, and Jeffrey Dean. 2013{\natexlab{a}}.
\newblock \href {https://arxiv.org/pdf/1301.3781.pdf} {{Efficient Estimation of
  Word Representations in Vector Space}}.
\newblock \emph{ICLR}.

\bibitem[{Mikolov et~al.(2013{\natexlab{b}})Mikolov, Sutskever, Chen, Corrado,
  and Dean}]{Mikolov2013}
Tomas Mikolov, Ilya Sutskever, Kai Chen, Greg Corrado, and Jeffrey Dean.
  2013{\natexlab{b}}.
\newblock \href
  {http://papers.nips.cc/paper/5021-distributed-representations-of-words-and-phrases-and-their-compositionality.pdf}
  {{Distributed Representations of Words and Phrases and their
  Compositionality}}.
\newblock \emph{NIPS}, pages 3111--3119.

\bibitem[{Mrk{\v{s}}ic et~al.(2016)Mrk{\v{s}}ic, S{\'{e}}aghdha, Thomson,
  Ga{\v{s}}iga{\v{s}}ic, Rojas-Barahona, Su, Vandyke, Wen, and
  Young}]{Mrksic2016}
Nikola Mrk{\v{s}}ic, Diarmuid~O S{\'{e}}aghdha, Blaise Thomson, Milica
  Ga{\v{s}}iga{\v{s}}ic, Lina Rojas-Barahona, Pei-Hao Su, David Vandyke,
  Tsung-Hsien Wen, and Steve Young. 2016.
\newblock \href {https://www.aclweb.org/anthology/N16-1018} {{Counter-fitting
  Word Vectors to Linguistic Constraints}}.
\newblock In \emph{Proceedings of NAACL-HLT 2016}, pages 142--148, San Diego,
  CA, USA. ACL.

\bibitem[{Mrk{\v{s}}ic et~al.(2017)Mrk{\v{s}}ic, Vulic, S{\'{e}}aghdha,
  Leviant, Reichart, Ga{\v{s}}ic, Korhonen, and Young}]{Mrksic2017}
Nikola Mrk{\v{s}}ic, Ivan Vulic, Diarmuid~{\'{O}} S{\'{e}}aghdha, Ira Leviant,
  Roi Reichart, Milica Ga{\v{s}}ic, Anna Korhonen, and Steve Young. 2017.
\newblock \href
  {https://www.mitpressjournals.org/doi/pdf/10.1162/tacl{\_}a{\_}00063}
  {{Semantic Specialization of Distributional Word Vector Spaces using
  Monolingual and Cross-Lingual Constraints}}.
\newblock \emph{Transactions of the Association for Computational Linguistics},
  5:309--324.

\bibitem[{Pedregosa et~al.(2011)Pedregosa, Varoquaux, Gramfort, Michel,
  Thirion, Grisel, Blondel, Prettenhofer, Weiss, Dubourg, Vanderplas, Passos,
  Cournapeau, Brucher, Perrot, and Duchesnay}]{Pedregosa2011}
Fabian Pedregosa, Ga{\"{e}}l Varoquaux, Alexandre Gramfort, Vincent Michel,
  Bertrand Thirion, Olivier Grisel, Mathieu Blondel, Peter Prettenhofer, Ron
  Weiss, Vincent Dubourg, Jake Vanderplas, Alexandre Passos, David Cournapeau,
  Matthieu Brucher, Matthieu Perrot, and {\'{E}}douard Duchesnay. 2011.
\newblock \href {http://www.jmlr.org/papers/v12/pedregosa11a} {{Scikit-learn:
  Machine Learning in Python}}.
\newblock \emph{Journal of Machine Learning Research}, 12:2825--2830.

\bibitem[{Pennington et~al.(2014)Pennington, Socher, and
  Manning}]{Pennington2014}
Jeffrey Pennington, Richard Socher, and Christopher~D Manning. 2014.
\newblock \href {https://nlp.stanford.edu/pubs/glove.pdf} {{GloVe: Global
  Vectors for Word Representation}}.
\newblock \emph{EMNLP}, pages 1532--1543.

\bibitem[{Rowley et~al.(1998)Rowley, Baluja, and Kanade}]{Rowley1998}
Henry~A Rowley, Shumeet Baluja, and Takeo Kanade. 1998.
\newblock \href
  {https://ieeexplore.ieee.org/stamp/stamp.jsp?tp={\&}arnumber=655647} {{Neural
  Network-Based Face Detection}}.
\newblock \emph{IEEE Transactions on Pattern Analysis and Machine
  Intelligence}, 20(1):23--38.

\bibitem[{Rubner et~al.(1998)Rubner, Tomasi, and Guibas}]{Rubner1998}
Yossi Rubner, Carlo Tomasi, and Leonidas~J Guibas. 1998.
\newblock \href {https://ieeexplore.ieee.org/stamp/stamp.jsp?arnumber=710701}
  {{A Metric for Distributions with Applications to Image Databases}}.
\newblock In \emph{Sixth International Conference on Computer Vision}, pages
  59--66.

\bibitem[{Schnabel et~al.(2015)Schnabel, Labutov, Mimno, and
  Joachims}]{Schnabel2015}
Tobias Schnabel, Igor Labutov, David Mimno, and Thorsten Joachims. 2015.
\newblock \href {https://doi.org/10.18653/v1/D15-1036} {{Evaluation methods for
  unsupervised word embeddings}}.
\newblock In \emph{Proceedings of the 2015 Conference on Empirical Methods in
  Natural Language Processing}.

\bibitem[{Sennrich et~al.(2016)Sennrich, Haddow, and Birch}]{Sennrich2016}
Rico Sennrich, Barry Haddow, and Alexandra Birch. 2016.
\newblock \href {http://arxiv.org/abs/1511.06709v4} {{Improving Neural Machine
  Translation Models with Monolingual Data}}.
\newblock \emph{arXiv}.

\bibitem[{Simard et~al.(2003)Simard, Steinkraus, and Platt}]{Simard2003}
Patrice~Y Simard, Dave Steinkraus, and John~C Platt. 2003.
\newblock \href
  {https://ieeexplore.ieee.org/stamp/stamp.jsp?tp={\&}arnumber=1227801} {{Best
  Practices for Convolutional Neural Networks Applied to Visual Document
  Analysis}}.
\newblock In \emph{Proceedings of the Seventh International Conference on
  Document Analysis and Recognition}, pages 958--963.

\bibitem[{Vulic et~al.(2018)Vulic, Glava{\v{s}}, Mrk{\v{s}}ic, and
  Korhonen}]{Vulic2018}
Ivan Vulic, Goran Glava{\v{s}}, Nikola Mrk{\v{s}}ic, and Anna Korhonen. 2018.
\newblock \href {https://www.aclweb.org/anthology/N18-1048.pdf}
  {{Post-Specialisation: Retrofitting Vectors of Words Unseen in Lexical
  Resources}}.
\newblock In \emph{Proceedings of NAACL-HLT 2018}, pages 516--527, New Orleans,
  LA, USA. ACL.

\bibitem[{Vulic and Mrk{\v{s}}ic(2018)}]{Vulic2018a}
Ivan Vulic and Nikola Mrk{\v{s}}ic. 2018.
\newblock \href {https://www.aclweb.org/anthology/N18-1103.pdf} {{Specialising
  Word Vectors for Lexical Entailment}}.
\newblock In \emph{Proceedings of NAACL-HLT 2018}, pages 1135--1145, New
  Orleans, LA, USA. ACL.

\bibitem[{Wong et~al.(2016)Wong, Gatt, Stamatescu, and McDonnell}]{Wong2016}
Sebastien~C Wong, Adam Gatt, Victor Stamatescu, and Mark~D McDonnell. 2016.
\newblock \href
  {https://ieeexplore.ieee.org/stamp/stamp.jsp?tp={\&}arnumber=7797091}
  {{Understanding data augmentation for classification: when to warp?}}
\newblock In \emph{2016 International Conference on Digital Image Computing:
  Techniques and Applications (DICTA)}, pages 1--6. IEEE.

\bibitem[{Xie et~al.(2019)Xie, Dai, Hovy, Luong, and Le}]{Xie2019}
Qizhe Xie, Zihang Dai, Eduard Hovy, Minh-Thang Luong, and Quoc~V Le. 2019.
\newblock \href {http://arxiv.org/abs/1904.12848v4} {{Unsupervised Data
  Augmentation for Consistency Training}}.
\newblock \emph{arXiv}.

\bibitem[{Yu and Dredze(2014)}]{Yu}
Mo~Yu and Mark Dredze. 2014.
\newblock \href {https://www.aclweb.org/anthology/P14-2089/} {{Improving
  Lexical Embeddings with Semantic Knowledge}}.
\newblock In \emph{Proceedings of the 52nd Annual Meeting of the Association
  for Computational Linguistics (Short Papers)}, pages 545--550, Baltimore, MD,
  USA. Association for Computational Linguistics.

\bibitem[{Zhang et~al.(2016)Zhang, Zhao, and Lecun}]{Zhang2016b}
Xiang Zhang, Junbo Zhao, and Yann Lecun. 2016.
\newblock \href {http://arxiv.org/abs/1502.01710} {{Character-level
  Convolutional Networks for Text Classification}}.
\newblock In \emph{Advances in Neural Information Processing Systems}, pages
  649--657. NeurIPS.

\end{thebibliography}
\bibliographystyle{acl_natbib}

\end{document}